\documentclass{article}

\usepackage{arxiv}

\usepackage[utf8]{inputenc} 
\usepackage[T1]{fontenc}    
\usepackage{url}            
\usepackage{booktabs}       
\usepackage{amsfonts}       
\usepackage{nicefrac}       
\usepackage{microtype}      
\usepackage{lipsum}
\usepackage{graphicx}
\usepackage{multirow}
\usepackage{booktabs}
\usepackage{enumitem}
\usepackage[labelsep=space]{caption}
\usepackage[table,xcdraw]{xcolor}
\usepackage[hidelinks]{hyperref}
\usepackage{algorithm}
\usepackage{algpseudocode,amsmath}
\usepackage{chngcntr}
\usepackage{wrapfig}
\usepackage{multicol}
\usepackage[export]{adjustbox}
\graphicspath{ {./images/} }

\makeatletter
\let\c@algorithm\c@figure
\renewcommand{\ALG@name}{Figure}

\renewcommand{\thefigure}{\textbf{\@arabic\c@figure}}
\makeatother

\newenvironment{Figure}
  {\par\medskip\noindent\minipage{\linewidth}}
  {\endminipage\par\medskip}

\newcommand{\var}{\texttt}
\newcommand{\assign}{\leftarrow}
\newcommand{\multilinestate}[1]{%
  \parbox[t]{\linewidth}{\raggedright\hangindent=\algorithmicindent\hangafter=1
    \strut#1\strut}}

\title{\Large{CONFLARE: CONFormal LArge language model REtrieval}}

\author{
\href{https://orcid.org/0000-0003-4664-0751}{\includegraphics[scale=0.06]{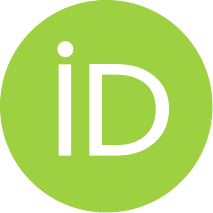}\hspace{1mm}Pouria Rouzrokh \scriptsize{\textnormal{MD, MPH, MHPE}}}$^\mathrm{1, 2, *}$ \\
   \And
   \href{https://orcid.org/0000-0003-3275-2971}{\includegraphics[scale=0.06]{orcid.pdf}\hspace{1mm}Shahriar Faghani \scriptsize{\textnormal{MD}}}$^\mathrm{1, *}$ \\
  \AND 
  \href{https://orcid.org/0009-0009-5139-4875}{\includegraphics[scale=0.06]{orcid.pdf}\hspace{1mm}Cooper Gamble$^\mathrm{1}$} \\
  \And
   \href{https://orcid.org/0000-0002-3940-8359}{\includegraphics[scale=0.06]{orcid.pdf}\hspace{1mm}Moein Shariatnia$^\mathrm{3}$} \\
  \AND 
  \href{https://orcid.org/0000-0001-7926-6095}{\includegraphics[scale=0.06]{orcid.pdf}\hspace{1mm}Bradley J. Erickson \scriptsize{\textnormal{MD, PhD}}}$^\mathrm{1, \dagger}$ \\
  \texttt{bje@mayo.edu} \\
}
\renewcommand{\headeright}{}
\renewcommand{\undertitle}{}
\renewcommand{\shorttitle}{}

\begin{document}
\pagestyle{plain}
\maketitle
\vspace{-20pt}
\begin{center}
	\footnotesize{(1) Mayo Clinic Artificial Intelligence Laboratory, Mayo Clinic, MN, USA, (2) Orthopedic Surgery Artificial Intelligence Laboratory, Mayo Clinic, MN, USA, (3) Tehran University of Medical Sciences, Tehran, Iran} \\
	\vspace{10pt}	
	\footnotesize{$^*$Co-first author, $^\dagger$Corresponding author} 

\end{center}
\vspace{16pt}
\begin{abstract}
	\hspace*{10pt}Retrieval-augmented generation (RAG) frameworks enable large language models (LLMs) to retrieve relevant information from a knowledge base and incorporate it into the context for generating responses. This mitigates hallucinations and allows updating the knowledge without retraining the LLM. However, RAG does not guarantee valid responses if retrieval fails to identify the necessary information as the context for response generation. Also, if there is contradictory content, the RAG response will likely reflect only one of the 2 possible responses. Therefore, quantifying uncertainty in the retrieval process is crucial for ensuring RAG trustworthiness. In this report, we introduce a four-step framework for applying conformal prediction to quantify retrieval uncertainty in RAG frameworks. First, a calibration set of questions answerable from the knowledge base is constructed. Each question's embedding is compared against document embeddings to identify the most relevant document chunks containing the answer and record their similarity scores. Given a user-specified error rate ($\alpha$), these similarity scores are then analyzed to determine a similarity score cutoff threshold. During inference, all chunks with similarity exceeding this threshold are retrieved to provide context to the LLM, ensuring the true answer is captured in the context with a ($1-\alpha$) confidence level. We provide a Python package that enables users to implement the entire workflow proposed in our work, only using LLMs and without human intervention.
\end{abstract}


\begin{center}
    \includegraphics[scale=0.03,valign=c]{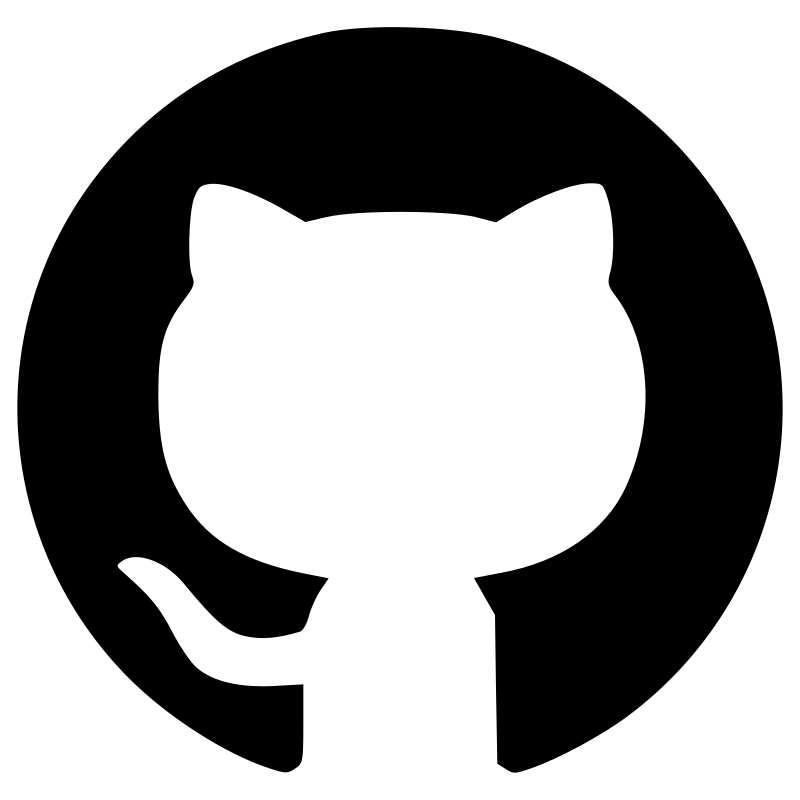}
    \raisebox{-0.3\height}{\parbox{0.6\textwidth}{\centering\href{https://github.com/Mayo-Radiology-Informatics-Lab/conflare}{\textbf{https://github.com/Mayo-Radiology-Informatics-Lab/conflare}}}}
\end{center}
\vspace{20pt}

\begin{multicols*}{2}
\section{Introduction}
\hspace*{10pt}Autoregressive large language models (LLMs) have recently attracted substantial interest, demonstrating capabilities that sometimes surpass human performance in diverse applications (1–3). However, despite their remarkable performances, LLMs are not without limitations. One of the most critical issues is their propensity for "hallucination"—the generation of incorrect responses or the citation of non-existent evidence with undue confidence (4). While such hallucinations might be tolerable in some contexts, they pose considerable challenges for the deployment of LLMs in fields such as medicine, where adherence to evidence-based practices is paramount. Moreover, the vast knowledge base of LLMs, while impressive, is essentially a snapshot of the publicly available online content at the time of their training. This static nature means that LLMs cannot automatically update their knowledge, a drawback that is particularly problematic in domains where access to the latest information is crucial.
\\
\hspace*{10pt}Retrieval-Augmented Generation (RAG) is a framework designed to overcome the two aforementioned limitations of LLMs (Figure 1) (5). Assuming a knowledge base containing the necessary information for generating a valid response to a given query, a RAG framework retrieves relevant pieces of information from the knowledge base, and then incorporates them into the context provided to the LLM for response generation. Consequently, the LLM leverages the provided information to answer the question accurately, reducing the likelihood of generating hallucinated content. Moreover, if the body of evidence relevant to the queried question has been updated since the LLM was last trained, it is only necessary to update the knowledge base rather than retrain the LLM. Due to its advantages and straightforward nature, RAG has emerged as a popular tool for enhancing the reliability of LLMs in question-answering applications. Furthermore, numerous variants of RAG have been proposed, seeking to refine the framework’s validity or efficiency.

\begin{figure*}
	\centering
	\frame{\includegraphics[width=\textwidth]{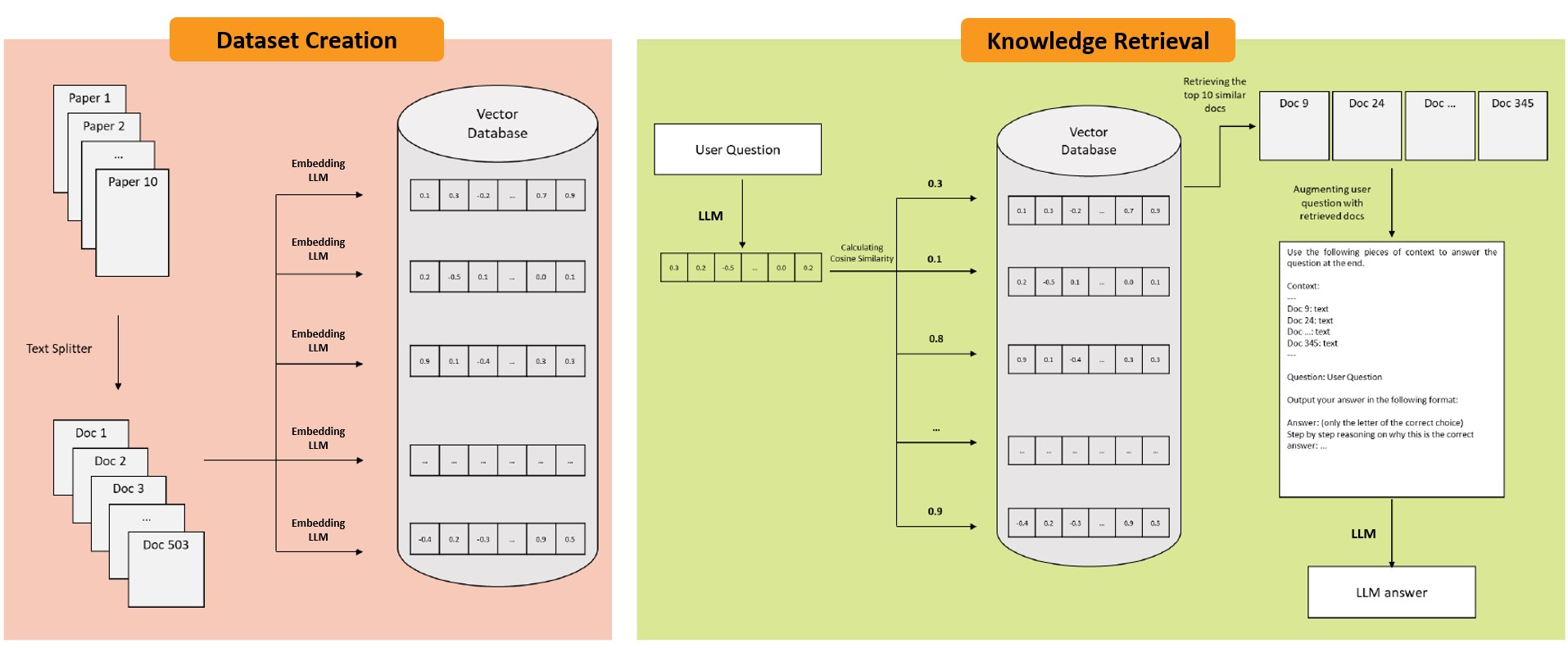}}
	\captionof{figure}{\centering The workflow of a conventional Retrieval Augmented Generation (RAG) framework. LLM = large language model.}
	\label{fig:fig1}
\end{figure*}

\hspace*{10pt}Despite the popularity of RAG, it does not guarantee that an LLM will generate a valid response to a query (6). This issue may arise due to uncertainties (ie. errors) associated with either the retrieval or the generation process. If the necessary information to answer the question does not exist in the knowledge base, or if the retrieval process fails to identify the most relevant information, the context provided to the LLM will lack the required details for accurately answering the question. Moreover, even with an appropriate context, an LLM might engage in erroneous reasoning based on that information or, worse, disregard the context entirely when generating a response. Thus, the application of RAG, like other machine learning (ML) applications, is subject to uncertainty. The greater the uncertainty of a model's response, the lower the trustworthiness of its outputs.
\\
\hspace*{10pt}On a positive note, tools exist that can quantify the uncertainty of most ML models, and recently, similar tools have been developed for quantifying the uncertainty of LLM generations. Most of these methods employ conformal prediction, a robust uncertainty quantification technique offering statistical guarantees about the reported uncertainty (7–9). However, to the best of our knowledge, no report exists on applying conformal prediction to the retrieval phase of an RAG-based pipeline. In this report, we introduce a straightforward and stepwise pipeline to quantify the uncertainty of the retrieval process in RAG, discussing its strengths and limitations. Additionally, we provide a Python package to automate this process using LLMs, along with a hands-on example notebook that demonstrates the package application to a medical question-and-answer scenario\footnote{https://github.com/Mayo-Radiology-Informatics-Lab/conflare}.

\section{Background}
\subsection{Knowledge Retrieval in RAG}
\hspace*{10pt}The core of a RAG framework involves a retrieval process that enables LLMs to access the most relevant information to a given quesiton for generating the required response. To establish this process, information must be stored in a vector database as a knowledge bank for the LLM, converting original textual information (or image or sound if the downstream model is multimodal), into mathematical representations for storage.
\\ 
\hspace*{10pt}The most conventional method for building such representations involves using embedding functions that compress the semantics of a piece of information into a vector (10). The function used to embed the data may be similar to or different from the downstream LLM employed for generating responses. Regardless, it is important to note that the embedding function, like the LLM itself, may also have inherent limitations, such as a predisposition to \textbf{uncertain performance}. Moreover, it is crucial to consider the length of text data that the embedding LLM will convert into a single vector. Feeding long text to that function results in a vector that abstracts concepts too generally or which may contain multiple concepts, which may hinder the retrieval process. Thus, it is common practice to divide available information into chunks of text of a certain length and then embed each chunk separately. However, chunking introduces additional nuances to the framework (e.g., determining the optimal chunk length, deciding how much overlap to include between chunks, etc.), but this is beyond the scope of this report (6).
\\
\hspace*{10pt}Given such a vector database, the next step in the RAG framework is to embed the input question using the same embedding function used to create the vector database. Then, the similarity between this question vector and all entries in the vector database can be computed to find those with the highest semantic similarity to the question vector. This semantic comparison can be formulated as a distance between each pair of vectors such as cosine similarity and dot product (11). 
\\
\hspace*{10pt}When the response vectors with the highest similarity (smallest distance) with respect to the question vector are identified, their corresponding chunks are retrieved and passed to the downstream LLM as a context of information upon which it should rely for generating responses. Note that the retrieval process has its own nuances. For example, what should be the distance threshold that separates relevant information from irrelevant information? If too many vectors seem relevant, how many of their corresponding chunks should be fed as the context to the downstream LLM? Moreover, how can one ensure that the downstream LLM will rely on the provided context to generate a response?

\subsection{Conformal Prediction}
\hspace*{10pt}Quantifying the uncertainty of DL models is crucial in the medical field, where outputs may influence patient care decisions (12). Conformal prediction, also known as conformal inference, introduces a straightforward approach to uncertainty quantification, providing uncertainty levels with statistical guarantees (13,14). Specifically, for any given input, conformal prediction estimates a prediction interval in regression tasks and a set of classes in classification tasks. These estimates are guaranteed to encompass the ground-truth value with a probability defined by the user, thus alerting the end-user to instances of unacceptably high uncertainty, which may require human verification of the model’s predictions.
\\
\hspace*{10pt}To implement conformal prediction in a supervised ML model, one must first compile a calibration set comprising input data points and their corresponding ground truth labels. Ideally, this calibration set should be from both the same data distribution but distinct from the validation set used for tuning the model’s hyperparameters, as well as from the unseen test set. Following this, the ML model is trained in the standard manner and then applied to the calibration data to generate predictions.
\\
\hspace*{10pt}With both ground truth and predicted labels from the calibration set, one can calibrate the conformal predictor. This process begins with specifying a maximum acceptable error level ($\alpha$) and developing a set of criteria to ensure the model’s predictions include the ground truth label within a probability of $1-\alpha$. These criteria, derived through mathematical operations on the calibration set, are then applicable to any subsequent inferences. Although various methods exist for calibrating a conformal predictor — each aiming to minimize the prediction set or range while maintaining the specified certainty level — detailing these techniques falls outside this discussion’s scope (15–18). We will later describe a simple calibration approach for the retrieval phase of RAG.
\\
\hspace*{10pt}A significant benefit of conformal prediction over other uncertainty quantification methods is its theoretical ability to provide a statistical guarantee over its prediction sets (13). Nonetheless, this claim presupposes that the calibration dataset accurately reflects the data distribution that the model will encounter in real-world applications.

\section{Conformal-Enhanced Retrieval}
\hspace*{10pt}One strategy for applying uncertainty quantification to any question-answering RAG framework involves enhancing the retrieval process with conformal prediction (Figure 2). This approach enables users to retrieve as many document chunks as necessary to ensure that the context provided to the downstream LLM contains the true answer to the user-specified question, with a level of uncertainty deemed tolerable by the user. The following is a stepwise roadmap to implement this strategy.
\begin{figure*}
	\centering
	\frame{\includegraphics[width=\textwidth]{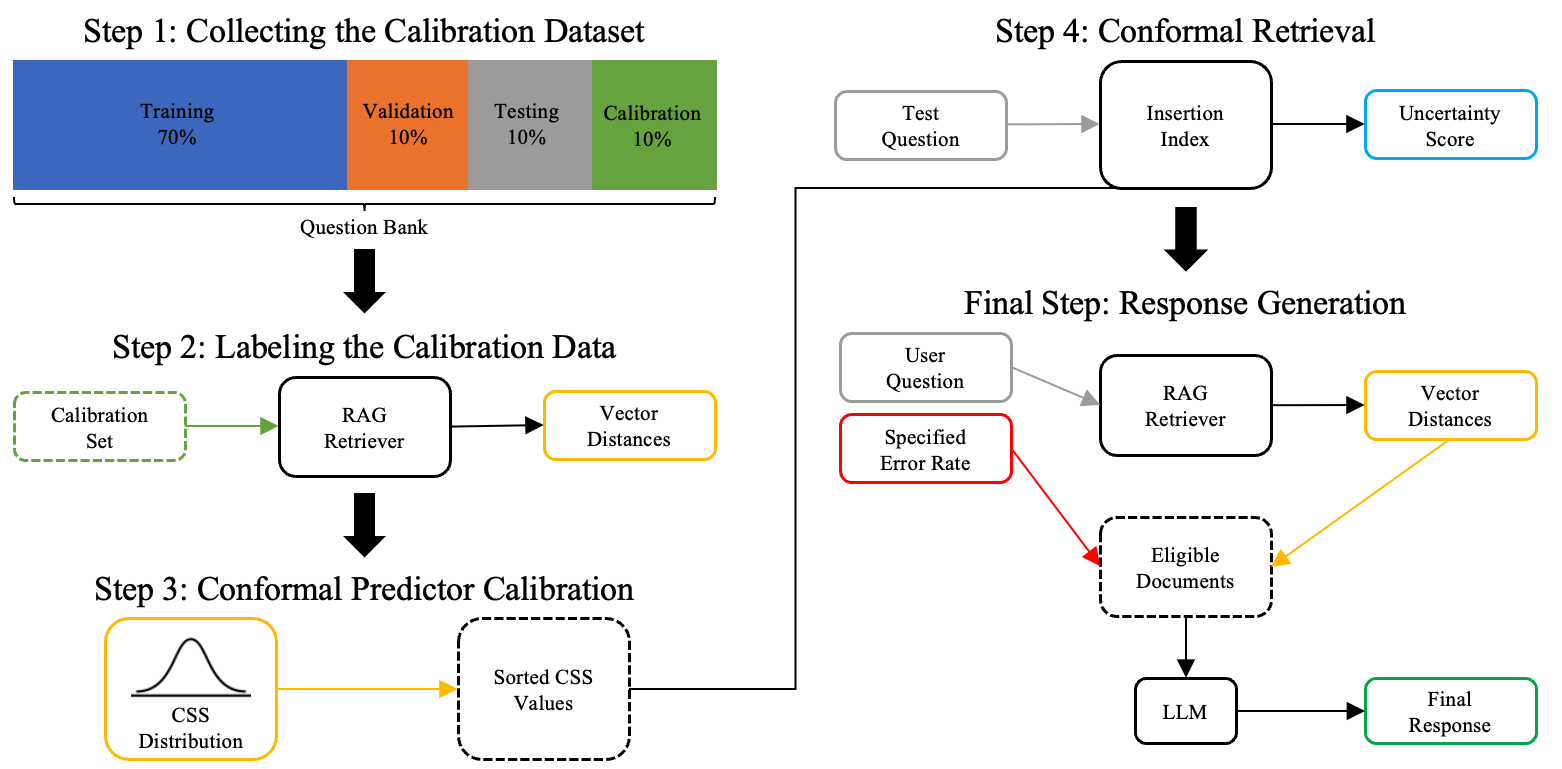}}
	\captionof{figure}{\centering Conformal retrieval framework from data collection to final response generation. CSS = cosine similarity score, LLM = large language model, RAG = retrieval-augmented generation.}
	\label{fig:fig1}
\end{figure*}

\subsection{Step 1: Collecting the Calibration Data}
To calibrate a retrieval process, input data should consist of questions that are answerable based on correct retrieval. When users have access to a comprehensive question-answering dataset for training (or fine-tuning) the LLM, they have two options: they can either allocate an entirely independent set of questions for calibration or, as previously mentioned, repurpose questions from the validation set for this role. The guidelines for splitting the calibration set mirror those for the validation/test sets: the questions in the calibration data must be representative of what is expected during inference time, and they must be unseen by the LLM during training.
\\
\hspace*{10pt}Enhancing the retrieval process with conformal prediction is even possible when end-users have no access to the datasets used for training or fine-tuning the model. Under such circumstances, users should craft a set of questions, each answerable based on at least one of the stored documents in the RAG vector database. The topics, quantity, and quality of the crafted questions should be similar to what the LLM is expected to encounter at inference. For instance, if a RAG framework is anticipated to be used for answering patients’ questions about their medications, one could craft several hundred questions on topics such as timing, dosage, or complications of various medications, written in non-technical language as though posed by real patients. It is crucial for each question to be grounded in (i.e., be answerable according to) at least one document accessible to the RAG vector database.
\\
\hspace*{10pt}Ideally, the questions described above should be created by humans who are representative of the target users of the RAG application. However, meeting such a requirement may not always be feasible. For example, recruiting humans could be costly, creating questions by users may take a significant amount of time, and, last but not least, a RAG application may be expected to work across multiple different knowledge domains, each requiring its own calibration set. To address these challenges, it is possible to use LLMs to create the questions. To do so, some reference documents are sampled from all the documents stored in the database, and an LLM is tasked with creating one question and answer from each document. Obviously, the higher the quality, relevancy, and diversity of these questions, the better the conformal predictor will be calibrated later in the process.

\subsection{Step 2: Labeling the Calibration Data}
\hspace*{10pt}After collecting the questions intended for use as calibration data, the next step is to identify the first document chunk within the RAG framework database that contains an answer to each question. This is achieved by first embedding all questions one by one, followed by performing the standard retrieval process and sorting all response vectors based on their similarity to the question vector. We then examine the document chunks corresponding to the response vectors, starting from the most relevant to the least relevant, to identify the first chunk that could potentially answer the question. It is important to note that this chunk might not be the original one used to generate the question, as the answer to some questions may be found in multiple documents. Upon identifying the appropriate document chunk for each question, we record the similarity between the response vector for that question and the question vector. These similarity values are collected and used as labels to calibrate the retrieval process conformal predictor.
\\
\hspace*{10pt}Please note that as with Step 1, the workflow described above could be done either manually or using an autonomous LLM agent. However, the task of evaluating a document chunk to see if it contains the response to a question may need specific knowledge or reasoning skills. Therefore, not all LLMs might perform appropriately for this task.

\subsection{Step 3: Conformal Predictor Calibration}
\hspace*{10pt}At this checkpoint, the end-user is required to specify an error rate percentage for the conformal prediction. For instance, if the end-user sets an error rate of 5\%, this implies that they expect the retrieval process to return a collection of chunks containing the correct answer to a queried question in 95\% or more of the inference runs. 
\\
\hspace*{10pt}By the end of Step 2, we should have a single similarity value for each question and response within the calibration data. As previously mentioned, this value indicates the similarity between the question vector and the most semantically similar response vector containing the answer to that question. We then sort these values from highest to lowest and identify the value of the list item that corresponds to the percentile determined by the user-specified error rate. This single similarity score will be our cutoff for deciding which document chunks should be returned from the conformal retrieval process in the next step. Figure 3 demonstrates a sample Python function that performs Step 3. 

\begin{Figure}
	\line(1,0){\textwidth}
	\begin{algorithmic}
		\scriptsize{\State \multilinestate{$\var{errorRate} \assign{}$user-specified value for the percentage of inference runs in which the correct answer to a queried question is present in the returned collection of chunks}
		\State \multilinestate{$\var{calibrationCSS} \assign{}$list of similarity scores for each question in the calibration dataset (from \textbf{Step 2})}
		\Function{performStepThree}{\var{errorRate, calibrationCSS}}
			\\ \hspace{20pt} \textbf{sort}(\var{calibrationCSS}, descending=\var{True})
			\\ \hspace{20pt} \var{cssThreshold} = \textbf{percentile}(calibrationCSS, $1.0-\text{errorRate}$)
			\\ \hspace{20pt} \textbf{return} \var{cssThreshold}
		\EndFunction}
	\end{algorithmic}
	\vspace{-7pt}
	\line(1,0){\textwidth}
	\captionof{figure}{Pseudocode for Step 3}
\end{Figure}

\subsection{Step 4: Conformal-Augmented Retrieval}
\hspace*{10pt}With the similarity cutoff score at hand, we can now perform a conformal-augmented retrieval for each question encountered during inference. To accomplish this, we should adhere to the standard retrieval workflow to determine the similarity scores between the question vector and all response vectors. In a traditional RAG framework, the user is expected to arbitrarily select a certain number of the most similar chunks to serve as the output of the retrieval process. However, in an augmented retrieval setup, we should return all chunks whose similarity to the question vector is higher than the cutoff score. This criterion ensures that the correct answer to the question is included within one or more of the returned chunks, with an error rate specified by the user. Figure 4 shows a pseudocode function that executes Step 4.
\\
\hspace*{10pt}Once the retrieval process concludes, the downstream LLM generates a response to the question in a manner similar to a traditional RAG framework: All retrieved chunks are amalgamated into the context, and the LLM is prompted to answer the queried question based solely on the provided context. It is important to note that typically, the error rate is set at values like 1\%, 5\%, or 10\%. However, users should be aware that lower error rates usually result in more conservative retrieval (lower cut-off scores) and an increased number of returned document chunks. Depending on the downstream LLM utilized for response generation, reducing the error rate beyond a certain point may exceed the context window limits or even impair its performance due to an overly lengthy input prompt, thereby undermining the utility of the entire RAG framework.

\begin{Figure}
	\line(1,0){\textwidth}
	\scriptsize{\begin{algorithmic}
		\State \multilinestate{$\var{cssThreshold} \assign{}$result of \textbf{Step 3}}
		\State \multilinestate{$\var{chunksAndCSS} \assign{}$dictionary of chunks retrieved during inference and their corresponding similarity scores}
		\Function{performStepFour}{\var{cssThreshold, chunksAndCSS}}
			\\ \hspace{20pt} \textbf{define }\var{chunksToReturn} = \var{[]}
			\\ \hspace{20pt} \textbf{for each} (chunk, css) \textbf{in} \var{chunksAndCSS}
			\\ \hspace{40pt} \textbf{if} \var{css} $>$ \var{cssThreshold} \textbf{then} 
			\\ \hspace{60pt} \textbf{insert} \var{chunk} \textbf{into} \var{chunksToReturn}
			\\ \hspace{20pt} \textbf{return} \var{chunksToReturn}
		\EndFunction
	\end{algorithmic}}
	\vspace{-7pt}
	\line(1,0){\textwidth}
	\captionof{figure}{Pseudocode for Step 4}
\end{Figure}

\section{Conclusion}
\hspace*{10pt}In conclusion, we proposed a \textit{post hoc}, framework-agnostic workflow to enhance the retrieval process within any RAG pipeline through the use of conformal prediction. More specifically, we addressed the issue that arbitrarily selecting a certain number of semantically relevant document chunks during the retrieval process can potentially act as a bottleneck for the entire RAG framework, resulting in a context that fails to encompass the true answer to the question. Instead, we demonstrated that the application of conformal prediction could yield a collection of retrieved chunks that, collectively, are guaranteed to include a valid response to the queried question, adhering to a level of error specified by the user. 
\\
\hspace*{10pt}Despite its potential benefits, it is crucial to recognize that the effectiveness of conformal-enhanced retrieval may be compromised by three main limitations: 
\\
\hspace*{10pt}First, the statistical guarantee of conformal prediction holds only if the calibration data is representative of the questions the framework is expected to encounter during inference. If the pool of generated questions is small, the quality of these questions significantly deviates from those at test time (for example, due to suboptimal question generation by an LLM), or if the document chunks containing valid answers to the calibration questions are not accurately identified, the reliability of conformal-enhanced retrieval becomes questionable. 
\\
\hspace*{10pt}Second, the effectiveness can be compromised if user-specified error rates are excessively high or if the embedding model performs suboptimally in distinguishing between different concepts of interest, potentially resulting in a set of retrieved document chunks too large to fit within the context window of downstream LLMs for response generation. 
\\
\hspace*{10pt}Lastly, the downstream LLM responsible for generating responses based on the provided context also faces uncertainty. Even a process with high retrieval confidence can lead to uncertain outcomes during response generation. A pertinent example of how uncertainty in the generation phase can impact overall RAG performance is observed when some retrieved documents offer contradictory information relevant to the query in question. In such scenarios, a precise retrieval process is expected to capture all pertinent documents, whether they are in agreement or contradiction. However, it becomes the duty of the downstream LLM to communicate this uncertainty to the user. Therefore, evaluating the uncertainty management capabilities of the downstream LLM is crucial for the development of a reliable RAG framework.
\end{multicols*}

\section{References}
\begin{enumerate}[label={(\arabic*)}]
    \item {
        Bojic L, Kovacevic P, Cabarkapa M. GPT-4 Surpassing Human Performance in Linguistic Pragmatics. arXiv [cs.CL]. 2023. \url{http://arxiv.org/abs/2312.09545}.
    }

    \item {
        Luo X, Rechardt A, Sun G, et al. Large language models surpass human experts in predicting neuroscience results. arXiv [q-bio.NC]. 2024. \url{http://arxiv.org/abs/2403.03230}.
    }

    \item {
        Kung TH, Cheatham M, Medenilla A, et al. Performance of ChatGPT on USMLE: Potential for AI-assisted medical education using large language models. PLOS Digit Health. 2023;2(2):e0000198.
    }

    \item {
        Huang L, Yu W, Ma W, et al. A Survey on Hallucination in Large Language Models: Principles, Taxonomy, Challenges, and Open Questions. arXiv [cs.CL]. 2023. \url{http://arxiv.org/abs/2311.05232}.
    }

    \item {
        Lewis P, Perez E, Piktus A, et al. Retrieval-augmented generation for knowledge-intensive NLP tasks. Adv Neural Inf Process Syst. 2020; abs/2005.11401. \url{https://proceedings.neurips.cc/paper/2020/hash/6b493230205f780e1bc26945df7481e5-Abstract.htm}.
    }

    \item {
        Barnett S, Kurniawan S, Thudumu S, Brannelly Z, Abdelrazek M. Seven Failure Points When Engineering a Retrieval Augmented Generation System. arXiv [cs.SE]. 2024. \url{http://arxiv.org/abs/2401.05856}.
    }

    \item {
        Kumar B, Lu C, Gupta G, et al. Conformal Prediction with Large Language Models for Multi-Choice Question Answering. arXiv [cs.CL]. 2023. \url{http://arxiv.org/abs/2305.18404}.
    }

    \item {
        Quach V, Fisch A, Schuster T, et al. Conformal Language Modeling. arXiv [cs.CL]. 2023. \url{http://arxiv.org/abs/2306.10193}.
    }

    \item {
        Kang M, Gürel NM, Yu N, Song D, Li B. C-RAG: Certified Generation Risks for Retrieval-Augmented Language Models. arXiv [cs.AI]. 2024. \url{http://arxiv.org/abs/2402.03181}.
    }

    \item {
        Wang L, Yang N, Huang X, Yang L, Majumder R, Wei F. Improving Text Embeddings with Large Language Models. arXiv [cs.CL]. 2023. \url{http://arxiv.org/abs/2401.00368}.
    }

    \item {
        Sitikhu P, Pahi K, Thapa P, Shakya S. A Comparison of Semantic Similarity Methods for Maximum Human Interpretability. arXiv [cs.IR]. 2019. \url{http://arxiv.org/abs/1910.09129}.
    }

    \item {
        Faghani S, Moassefi M, Rouzrokh P, et al. Quantifying Uncertainty in Deep Learning of Radiologic Images. Radiology. Radiological Society of North America; 2023;308(2):e222217.
    }

    \item {
        Angelopoulos AN, Bates S. A Gentle Introduction to Conformal Prediction and Distribution-Free Uncertainty Quantification. arXiv [cs.LG]. 2021. \url{http://arxiv.org/abs/2107.07511}.
    }

    \item {
        Faghani S, Gamble C, Erickson BJ. Uncover This Tech Term: Uncertainty Quantification for Deep Learning. Korean J Radiol. 2024;25(4):395–398.
    }

    \item {
        Fontana M, Zeni G, Vantini S. Conformal prediction: A unified review of theory and new challenges. BJOG. Bernoulli Society for Mathematical Statistics and Probability; 2023;29(1):1–23.
    }

    \item {
        Boström H, Johansson U, Löfström T. Mondrian conformal predictive distributions. In: Carlsson L, Luo Z, Cherubin G, An Nguyen K, editors. Proceedings of the Tenth Symposium on Conformal and Probabilistic Prediction and Applications. PMLR; 08--10 Sep 2021. p. 24–38.
    }

    \item {
        Angelopoulos AN, Bates S, Fisch A, Lei L, Schuster T. Conformal Risk Control. arXiv [stat.ME]. 2022. \url{http://arxiv.org/abs/2208.02814}.
    }

    \item {
        Sun J, Carlsson L, Ahlberg E, Norinder U, Engkvist O, Chen H. Applying Mondrian Cross-Conformal Prediction To Estimate Prediction Confidence on Large Imbalanced Bioactivity Data Sets. J Chem Inf Model. 2017;57(7):1591–1598.
    }
\end{enumerate}
\end{document}